%% file: main.tex
\documentclass[acmtog, authorversion, nonacm]{acmart}

\usepackage{booktabs} 
\usepackage{multirow}
\usepackage{xspace}

\usepackage[ruled]{algorithm2e} 

\SetAlFnt{\small}
\SetAlCapFnt{\small}
\SetAlCapNameFnt{\small}
\SetAlCapHSkip{0pt}

\setcopyright{none}
\renewcommand\footnotetextcopyrightpermission[1]{}

\makeatletter
\fancypagestyle{standardpagestyle}{%
  \fancyhf{}%
}
\fancypagestyle{firstpagestyle}{%
  \fancyhf{}%
}
\makeatother
\begin{document}
\title{Compression and Retrieval: Implicit Memory Retrieval for Video World Models}

\author{Zhan Peng}
\authornote{Work done during an internship at HUJING Digital Media \& Entertainment Group}
\affiliation{%
 \institution{Huazhong University of Science and Technology}
 \country{China}}

\author{Jie Ma}
\affiliation{%
 \institution{HUJING Digital Media \& Entertainment Group}
 \country{China}}

\author{Huiqiang Sun}
\affiliation{%
 \institution{Huazhong University of Science and Technology}
 \country{China}}

\author{Chong Gao}
\authornotemark[1]
\affiliation{%
 \institution{Sun Yat-sen University}
 \country{China}}

\author{Zhijie Xue}
\author{Zhiyu Pan}
\author{Zhiguo Cao}
\authornote{Corresponding authors}
\affiliation{%
 \institution{Huazhong University of Science and Technology}
 \country{China}}

\author{Jun Liang}
\authornotemark[2]
\author{Jing Li}
\affiliation{%
 \institution{HUJING Digital Media \& Entertainment Group}
 \country{China}}

\newcommand{\name}{CaR\xspace}

\begin{teaserfigure}
    \centering
    \includegraphics[width=\linewidth]{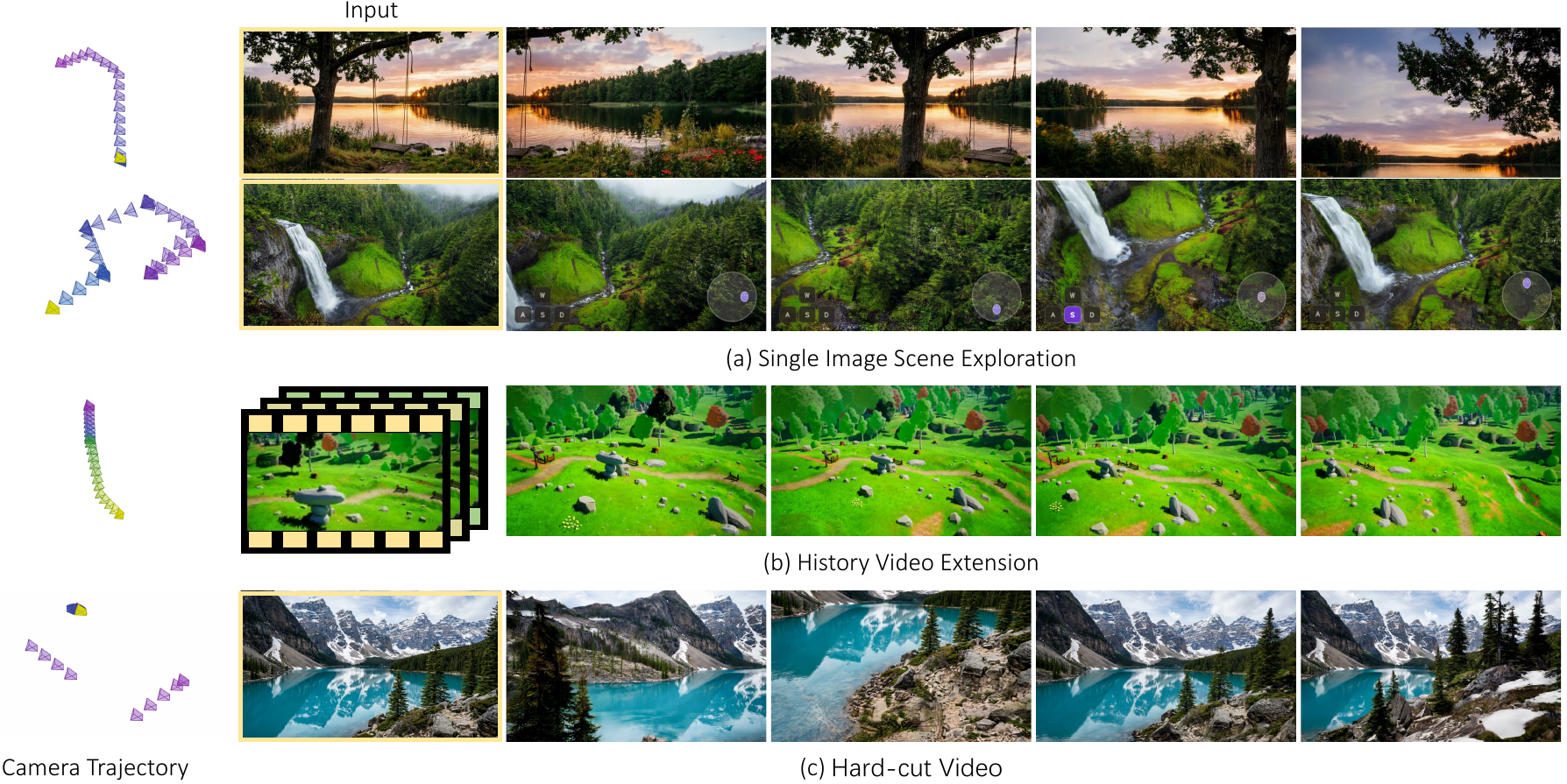}
    \captionof{figure}{
    \textbf{Teaser.} We propose \textbf{Compression and Retrieval}, an attention-driven implicit memory retrieval mechanism that operates flexibly and globally across the historical context. This figure illustrates three key applications enabled by our model's memory capabilities. Beyond performing single-image scene exploration, our approach also supports video-to-video generation while maintaining scene consistency. Notably, our method uniquely facilitates the synthesis of hard-cut videos, where the generated camera trajectories are discontinuous relative to the input context. In the visualization, blue paths represent the context trajectories, while red segments denote the trajectories of the generated videos. As demonstrated, our method consistently preserves scene consistency across all three settings, showcasing exceptional memory retrieval performance and precise control over complex camera trajectories.
    } 
    \label{fig:teaser}
\end{teaserfigure}

\maketitle

\input{section/abstract}
\input{section/introduction}
\input{section/related_work}
\input{section/method}

\input{section/experiments}

\input{section/conclusion}

\bibliographystyle{ACM-Reference-Format}
\bibliography{bibliography}

\appendix
\input{section/appendix}

\end{document}

%% file: section/abstract.tex
Video world models hold promise for simulating interactive environments, yet maintaining consistent long-term memory across complex camera trajectories remains a critical challenge. Existing methods typically rely on computationally expensive context scaling or rigid heuristic retrieval mechanisms, which lacks generalization to varying camera trajectories and environments. In this paper, we propose Compression and Retrieval~(\name), an attention-driven implicit memory retrieval mechanism to overcome these limitations. By injecting viewpoint information via positional encoding, our method performs flexible memory retrieval through attention computation. To efficiently process extended contexts with minimal computational overhead, we further introduce a lightweight context compression network. Furthermore, we construct SceneFly, a large-scale synthetic dataset featuring realistic camera trajectories and frame-level annotations to train and evaluate long-horizon video world models. Extensive experiments demonstrate that our approach achieves state-of-the-art results on established benchmarks and exhibits strong generalization to open-domain scenes. Project page: \url{https://github.com/Orange-3DV-Team/CaR}

%% file: section/introduction.tex
\begin{figure*}[]
    \centering
    \includegraphics[width=\linewidth]{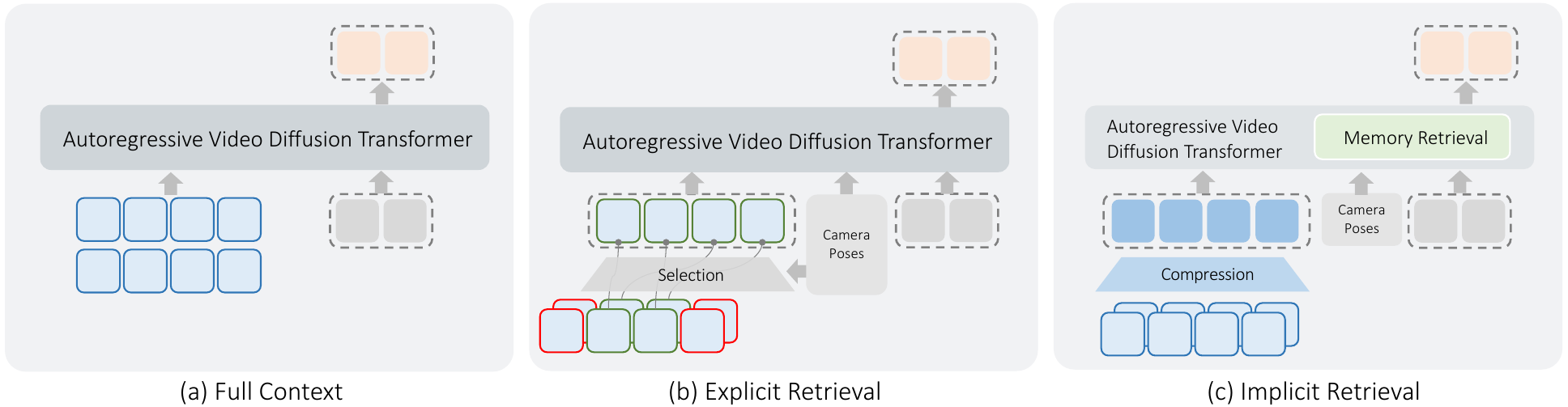}

    \captionof{figure}{
    \textbf{Comparison of Memory Paradigms.} (1) Scaling up the context window to utilize the entire context as memory is computationally prohibitive, rendering it impractical. (2) Explicit retrieval relies on hand-crafted heuristic rules; the rigidity of these rules severely restricts the model's generalization across diverse camera trajectories and scenes. (3) In contrast, our implicit retrieval is an attention-driven mechanism where the model performs retrieval directly within the global context, yielding both greater flexibility and superior performance.
    } 
    \label{fig:motivation}

\end{figure*}
\section{Introduction}
\label{sec:intro}

Diffusion-based video generation~\cite{hong2022cogvideo,kong2024hunyuanvideo,yang2024cogvideox,wan2025,hunyuanvideo2025} is evolving into interactive world models capable of simulating complex visual environments~\cite{hong2025relic,lingbot-world}, offering significant potential for long-horizon applications in autonomous driving~\cite{hu2023gaia,gao2024vista} and embodied intelligence~\cite{yang2023learning,wu2026pragmatic}. However, serving as reliable simulators for these tasks introduces a fundamental challenge: the model must maintain a robust long-term memory of the environment. Specifically, when previously observed regions re-enter the field of view after complex camera trajectories, the model must render them again consistently, even from entirely novel viewpoints.

Existing approaches expose a fundamental trade-off between memory coverage and efficiency. Full-context models~\cite{hong2025relic,lingbot-world} retain all recent frames, but the quadratic cost of temporal attention makes long histories increasingly impractical, as shown in Fig.\ref{fig:motivation}(a). Retrieval-based models reduce this cost by selecting a small subset of history according to heuristic rules, such as field-of-view (FoV)~\cite{yu2025context} or feature similarity~\cite{chen2026hydra}, as shown in Fig.\ref{fig:motivation}(b). These explicit retrieval rules are efficient when the target resembles a stored view, but they separate retrieval from generation and impose a fixed notion of relevance. Consequently, they can discard useful evidence under large viewpoint changes, discontinuous trajectories, or scene-dependent occlusions. The central challenge is therefore to preserve access to the full history while allowing the generation model itself to determine which memory is relevant.

To overcome these limitations, we propose \name, an attention-driven implicit memory retrieval mechanism that operates flexibly and globally across the historical context. Rather than selecting frames before generation, \name concatenates compressed context tokens with the target video tokens and retrieves memory through attention. We encode each token with its camera projection matrix so that context--target interactions depend on relative viewpoint geometry. A zero-initialized Retrieval Attention branch learns this geometry-aware interaction in parallel with the pretrained self-attention branch, preserving the base model while adding camera control and long-term retrieval. Because global attention over uncompressed history remains expensive, a dual-branch context encoder combines low-resolution scene structure with details extracted from high-resolution VAE latents, reducing the number of context tokens by approximately 97\%.

Training such a model requires trajectories that repeatedly revisit scene regions with accurate camera poses. Existing video collections provide limited revisiting behavior or noisy pose estimates, so we construct SceneFly using Unreal Engine~5. SceneFly contains diverse indoor, outdoor, and stylized environments, stochastic long-range camera trajectories, and exact frame-level camera annotations. Training on SceneFly and the real-world SpatialVid dataset, \name outperforms full-context and explicit-retrieval baselines on video extension and scene revisiting. It also supports hard cuts between discontinuous viewpoints, a setting that removes local frame continuity and directly tests retrieval from long-term memory.
 
Our contributions are threefold:
\begin{itemize}
   \item We propose Compression and Retrieval, an attention-driven implicit memory retrieval mechanism for scene-consistent interactive video generation.
   \item We introduce SceneFly, a large-scale revisiting synthetic dataset with frame-level camera annotations designed for training and evaluating memory capability.
   \item Our method achieves state-of-the-art results on established benchmarks and demonstrates strong generalization to open-domain scenes.
\end{itemize}

%% file: section/related_work.tex
\section{Related Work}

\paragraph{Long Video Generation.}

Modern video diffusion models~\cite{kong2024hunyuanvideo,yang2024cogvideox,brooks2024video,wan2025,hunyuanvideo2025} commonly combine Diffusion Transformers (DiTs)~\cite{peebles2023scalable} with rectified flow~\cite{lipman2023flow,liu2023flow,esser2024scaling}. However, their temporal attention cost limits the duration that can be jointly generated. Autoregressive methods extend videos segment by segment~\cite{henschel2024streamingt2v,chen2024seine,lu2024freelong,guo2025lct}, while context-compression methods summarize earlier segments~\cite{zhang2026frame,zhang2025pretraining,chen2026contextforcing}. These techniques improve duration but primarily rely on text as control and do not address the retrieval of scene content along prescribed camera trajectories.

\paragraph{Camera-Controlled Video Generation.}

Integrating camera parameters into video generation enables fine-grained viewpoint control~\cite{guo2024animatediff,baisyncammaster,liu2025free4d}. Early work~\cite{liu2023zero123} establishes the viability of camera-conditioned generation via zero-shot novel view synthesis, and subsequent methods encode camera poses through dedicated modules~\cite{wang2024motionctrl,bai2025recammaster,yu2025trajectorycrafter,he2025cameractrl,zheng2024cami2v,yu2025viewcrafter} or construct 3D geometric caches~\cite{ren2025gen3c} to scaffold world-consistent generation. PRoPE~\cite{li2025prope} embeds camera projection matrices into attention, and UCPE~\cite{zhang2025ucpe} generalizes this relative encoding across DiT blocks. Combined with causal generation~\cite{yin2025causvid,huang2026selfforcing}, these models can follow long camera paths, but content outside the active causal window is no longer directly accessible. We build on relative camera encoding for a different purpose: retrieving compressed observations from the complete history when old viewpoints become relevant again.

\paragraph{Video World Models.}

Interactive world models~\cite{bruce2024genie,alonso2024diamond,valevski2024gamengen,agarwal2025cosmos,yu2025gamefactory,hunyuanworld2025tencent,li2025hunyuangamecrafthighdynamicinteractivegame,hyworld22026} require both camera controllability and long-term historical memory. Lingbot-World~\cite{lingbot-world} and RELIC~\cite{hong2025relic} expand the temporal context, whereas Context-as-Memory~\cite{yu2025context} retrieves frames using field-of-view overlap and HyDRA~\cite{chen2026hydra} uses feature-similarity ranking. These methods adopt \emph{explicit retrieval}: fixed, hand-crafted criteria decide which frames to retrieve, making the selection opaque to the generation model. Geometry-based methods~\cite{li2025vmem,wu2025longtermvwm} instead store observations in explicit 3D structures, making performance dependent on reconstruction quality. In contrast, \name introduces \emph{implicit memory retrieval}, which retains a compressed global history and lets geometry-aware attention learn a soft, token-level retrieval policy jointly with video generation.

%% file: section/method.tex
\section{Method}
\begin{figure*}[t]
    \centering
    \includegraphics[width=\linewidth]{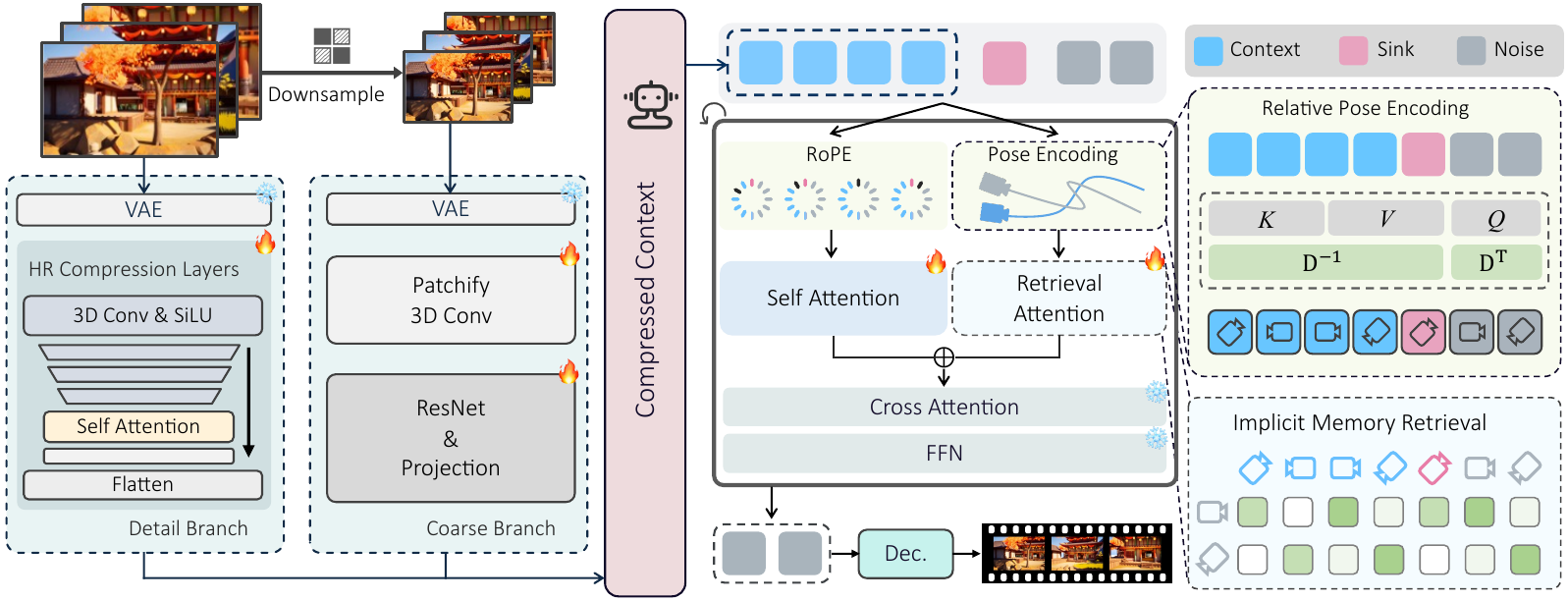}
    \vspace{-12pt}
    \captionof{figure}{
    \textbf{Overview of \name.}~A dual-branch compression network converts the historical video into compact context tokens. The context, an uncompressed sink frame, and noisy target tokens are then processed by two parallel attention branches: standard self-attention preserves the pretrained video prior, while Retrieval Attention uses relative camera poses to retrieve relevant history and control the target viewpoint.} 
    \label{fig:pipeline}

\vspace{-10pt}
\end{figure*}

We propose \name, an implicit memory retrieval mechanism. In this section, we describe the implementation of our approach.
Sec.~\ref{sec:prelim} introduces the preliminaries related to video generation. Sec.~\ref{sec:implicit} details the Implicit Memory Retrieval mechanism, focusing on how viewpoint information is injected into the latents and how global retrieval is performed within the context. Sec.~\ref{sec:context_compression} describes a learnable method to compress the context more compactly to save resources. Finally, Sec.~\ref{sec:datapipeline} introduces our long-video dataset collected using Unreal Engine 5, which features precise camera pose annotations, diverse scenes, and caption annotations.

\subsection{Preliminaries}
\label{sec:prelim}

\noindent\textbf{Video Diffusion Model.}
\name builds upon a pretrained text-to-video latent diffusion model formulated within the Rectified Flow framework~\cite{esser2024scaling}. Given an input video $V_0$, a video variational autoencoder (VAE) first encodes it into a clean latent representation $z_0$. A noisy latent $z_t$ is subsequently derived by linearly interpolating between $z_0$ and standard Gaussian noise $\epsilon \sim \mathcal{N}(0,1)$ at a randomly sampled timestep $t$:
\begin{equation}z_t = (1-t)z_0 + t\epsilon.
\end{equation}
During training, the noisy latent $z_t$ is processed by a Transformer-based diffusion model (DiT)~\cite{peebles2023scalable}, which predicts the instantaneous velocity field $v_{\Theta}(z_t, t)$ governing the ordinary differential equation (ODE) $\mathrm{d}z_t = v_{\Theta}(z_t, t)\,\mathrm{d}t$. The network is optimized using the Flow Matching objective:
\begin{equation}
\mathcal{L}_{FM} = \mathbb{E}_{z_0,t,\epsilon} || v_{\Theta}(z_t, t) - u_t(z_0 | \epsilon)||_2,
\end{equation}
where $u_t(z_0 \mid \epsilon)$ denotes the target velocity vector connecting $z_0$ and $\epsilon$. At inference, the generation process is initialized with pure noise $z_1$ and sequentially solved via Euler integration along a decreasing timestep schedule $\{t_i\}$, with $t_0 = 1$ and $t_n = 0$:
\begin{equation}
z_{t_{i+1}} = z_{t_i} + v_{\Theta}(z_{t_i}, t_i) \cdot (t_{i+1} - t_i).
\end{equation}Finally, the estimated latent $z_0$ is decoded back into the pixel space by the VAE to synthesize the output video.

\noindent\textbf{Camera Representation.}
Each frame's camera is parameterized by extrinsics $(\mathbf{R}, \mathbf{t}) \in \text{SE}(3)$ and shared intrinsics $\mathbf{K} \in \mathbb{R}^{3 \times 3}$.
We compose these into a $4 \times 4$ projective matrix:
\begin{equation}
    \mathbf{P} =
    \begin{bmatrix}
        \mathbf{K}\mathbf{R} & \mathbf{K}\mathbf{t} \\
        \mathbf{0}^\top & 1
    \end{bmatrix}
    \in \mathbb{R}^{4 \times 4}.
\end{equation}
This matrix is the transformation for computing camera coordinates from world coordinates.

\subsection{Implicit Memory Retrieval}
\label{sec:implicit}

\noindent\textbf{Motivation.} 
As analyzed in Sec.~\ref{sec:intro}, explicit memory retrieval relies on hand-crafted filtering rules, which restricts the model's flexibility and degrades its generalization across various trajectories and scenes. 
To eliminate this reliance on manual rules, we aim to enable the model to learn global implicit retrieval within the full context. Specifically, we enable the model to utilize the attention mechanism for retrieval, which necessitates accounting for spatial relevance in addition to semantic correlations.
Consequently, effective implicit retrieval necessitates assigning a camera pose-based "identity" to the features, thereby equipping the model with essential spatial awareness. 
Inspired by PRoPE~\cite{li2025prope} and UCPE~\cite{zhang2025ucpe}, we employ a relative encoding mechanism to capture relative spatial relationships, thereby enhancing the model's implicit retrieval capabilities.

\noindent\textbf{Relative Pose Encoding.}
For each token $i$ with $d$ channels, we construct a per-token transformation matrix for half of its channels from its camera's projective matrix $\mathbf{P}_{i}$ to replace the standard RoPE:
\begin{equation}
    \mathbf{D}_i = \mathbf{I}_{d/8} \otimes \mathbf{P}_{i} \in \mathbb{R}^{d/2 \times d/2},
\end{equation}
where $\otimes$ means Kronecker product and $I_{d/8}$ is an identity matrix.
We apply $\mathbf{D}_i$ to queries and its inverse $\mathbf{D}_i^{-1}$ to keys and values via batched matrix-vector products, yielding the attention:
\begin{equation}
    O = \mathbf{D} \odot \text{Attn}\left( \mathbf{D}^{\top} \odot Q,~\mathbf{D}^{-1} \odot K,~\mathbf{D}^{-1} \odot V \right),
\end{equation}
where $\odot$ denotes the batched matrix-vector product.
The key insight is that each dot product in the attention score naturally resolves to a relative projective transformation:
\begin{equation}
    \mathbf{D}_{i} \mathbf{D}_{j}^{-1} = \mathbf{I}_{d/8} \otimes \left( \mathbf{P}_{i} \mathbf{P}_{j}^{-1} \right),
\end{equation}

When two viewpoints are close, the relative projective transformation $\mathbf{P}_i \mathbf{P}_j^{-1}$ approaches the identity matrix, reducing to standard attention. Conversely, when there is a large deviation between the two viewpoints, the relative projective transformation exerts a suppressive effect, thereby diminishing the attention scores.

\noindent\textbf{Memory Retrieval.}
Leveraging the properties of relative pose encoding, we enable the model to perform retrieval directly within the attention mechanism, called Retrieval Attention. Specifically, the attention scores of latents irrelevant to the current viewpoint are suppressed, whereas those of highly relevant latents are relatively amplified. This naturally up-weights features from geometrically similar views, as illustrated in the Fig.~\ref{fig:pipeline}. Consequently, this property effectively filters features from nearby viewpoints without requiring any explicit frame selection or hand-designed retrieval criteria.

To preserve the inherent capabilities of the base model, we introduce a zero-initialized Retrieval Attention branch in parallel with the standard self-attention mechanism. The original self-attention retains the standard RoPE, while the Retrieval Attention employs the relative pose encoding. This dual-branch architecture ensures that implicit memory retrieval and camera control capabilities are integrated into the model without compromising the performance of the base model.

In general, implicit memory retrieval leverages relative pose encoding to inject camera poses directly into the attention operation, executing the retrieval process via an independent Retrieval Attention module. Our paradigm obviates the need for explicit frame selection or hand-crafted retrieval criteria, thereby promoting greater flexibility and generalization to novel camera trajectories and diverse scenes.

\begin{table*}[!ht]
\centering
\small

\caption{\textbf{Quantitative Comparison Results.}~Empowered by the Implicit Memory Retrieval mechanism, our model achieves the best memory capabilities and the highest visual quality. Conversely, the suboptimal results of CaM and HyDRA highlight the inherent rigidity of Explicit Retrieval methods that rely on hand-crafted heuristics. Furthermore, the performance degradation observed with naive downsampling~(Ours-DS) confirms the necessity of our Context Compression module. Finally, our favorable comparisons against the heavily parameterized LingBot-World further underscore the superiority of our approach.}
\vspace{-10pt}
\label{tab:main}
\setlength{\tabcolsep}{2.5pt}
\begin{tabular}{llcccccccc}
\toprule
\multirow{2}{*}{Dataset} & \multirow{2}{*}{Method} & \multicolumn{4}{c}{Video Extension Comparison} & \multicolumn{4}{c}{Scene Revisiting Comparison} \\
\cmidrule(lr){3-6} \cmidrule(lr){7-10}
& & PSNR$\uparrow$ & SSIM$\uparrow$ & LPIPS$\downarrow$ & FVD$\downarrow$ & PSNR$\uparrow$ & SSIM$\uparrow$ & LPIPS$\downarrow$ & FVD$\downarrow$ \\
\midrule
\multirow{5}{*}{SceneFly}
& Lingbot~\cite{lingbot-world} & 16.37 & 0.516 & 0.450 & 143.2 & 15.63 & 0.470 & 0.499 & 87.2 \\
& CaM~\cite{yu2025context} & 21.45 & 0.627 & 0.171 & 53.1 & 19.39 & 0.584 & 0.271 & 46.7 \\
& HyDRA~\cite{chen2026hydra} & 21.25 & 0.620 & 0.176 & 54.2 & 20.38 & 0.610 & 0.235 & 43.3 \\
& Ours-DS & 19.56 & 0.561 & 0.246 & 149.6 & 18.03 & 0.527 & 0.345 & 274.6 \\
& Ours & \textbf{22.91} & \textbf{0.693} & \textbf{0.140} & \textbf{52.7} & \textbf{21.23} & \textbf{0.672} & \textbf{0.209} & \textbf{42.1} \\
\midrule
\multirow{5}{*}{SpatialVid}

& Lingbot~\cite{lingbot-world} & 17.59 & 0.577 & 0.342 & 122.5 & 16.11 & 0.568 & 0.360 & 93.7 \\
& CaM~\cite{yu2025context} & 18.00 & 0.565 & 0.278 & 94.9 & 18.88 & 0.621 & 0.292 & 50.6 \\
& HyDRA~\cite{chen2026hydra} & 17.81 & 0.560 & 0.291 & 105.7 & 19.90 & 0.648 & 0.244 & 44.3 \\
& Ours-DS & 17.22 & 0.543 & 0.347 & 169.6 & 16.07 & 0.555 & 0.429 & 537.3 \\
& Ours & \textbf{18.17} & \textbf{0.593} & \textbf{0.271} & \textbf{92.6} & \textbf{20.77} & \textbf{0.699} & \textbf{0.231} & \textbf{41.2} \\
\bottomrule
\vspace{-14pt}
\end{tabular}
\end{table*}

\subsection{Context Compression}
\label{sec:context_compression}

Implicit retrieval fundamentally changes the role of context compression. Explicit methods encode only frames selected by a predefined rule, whereas \name preserves the complete history and lets the generator determine relevance for each target query. Keeping every historical latent at its original resolution, however, would reintroduce the quadratic cost of full-context attention. Compression must therefore make global access affordable without prematurely deciding which observations will matter to a future, unknown target view. It should remove redundancy while preserving the generator's freedom to retrieve.

This objective differs from conventional video compression, which reconstructs each input frame. Our compressed history instead serves as memory for generation across potentially large viewpoint changes. It must retain global layout and coarse geometry to locate observed regions, as well as textures, boundaries, and thin structures to reproduce their appearance. Aggressive pixel-space downsampling preserves structure but removes fine details, while compressing only high-resolution VAE latents places a greater burden on the learnable encoder. A single representation therefore struggles to preserve both forms of evidence within a compact token budget.

We address this trade-off with two complementary paths. The coarse branch downsamples the context by $(2,4,4)$ before applying the frozen VAE, efficiently preserving scene layout and long-range structure. The detail branch first VAE-encodes the original-resolution context and then applies a lightweight latent downsampler, retaining appearance cues lost by pixel-space compression. Their outputs form a compact memory sequence containing both structural and fine-grained evidence.

Relative to the latent tokens before Context Compression, this configuration reduces the context token count by approximately 97\%, decoupling historical coverage from attention cost. Ours-DS in Table~\ref{tab:main} shows that compression rate alone is insufficient: naive downsampling reaches a similar token budget but discards retrieval-relevant evidence, whereas the dual-branch memory better reconstructs revisited content.

\subsection{Data Pipeline}
\label{sec:datapipeline}

Long-term memory requires more than long videos: training must revisit previously observed regions from new viewpoints while presenting both relevant and distracting history. Real videos~\cite{wang2025spatialvid} offer visual diversity but typically lack reliable camera poses and controlled revisit trajectories. Existing synthetic data~\cite{yu2025context} provides accurate poses, yet often has limited scene or motion diversity and insufficient variation in target-context relevance. These limitations allow models to exploit short-range continuity instead of learning retrieval.

SceneFly is designed for implicit retrieval. It contains approximately 1,000 minutes of video from 100 diverse indoor, outdoor, and stylized UE5 scenes, with exact frame-level intrinsics and extrinsics. Independently composed camera motions create long trajectories with substantial viewpoint changes and revisits. Each target is paired with context clips spanning strong, partial, and no FoV overlap: overlapping views provide reusable evidence, partial overlap requires geometric reasoning, and disjoint views serve as distractors. Varying the context length exposes the model to different retrieval difficulties without explicitly labeling which frame to use.

We collect trajectories through automated NavMesh exploration, construct target--context pairs by graded FoV overlap, and caption only target appearance with Qwen3-VL~\cite{Qwen3-VL}, excluding camera-motion descriptions. Please refer to Appendix~\ref{sec:appendix_scenefly} for full construction details.

%% file: section/experiments.tex
\section{Experiments}
\label{exp}

\begin{figure*}[t]
    \centering
    \includegraphics[width=\linewidth]{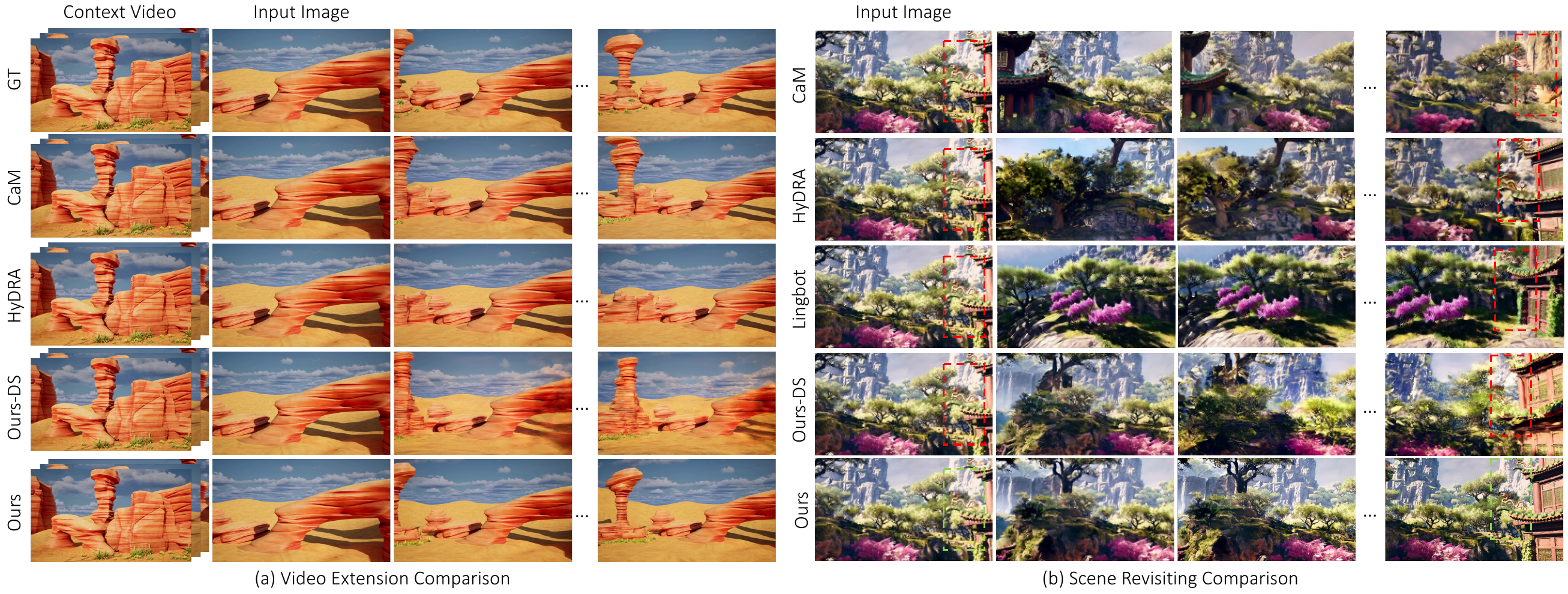}
    \vspace{-19pt}
    \captionof{figure}{
    \textbf{Qualitative Comparison Results.}~Compared to existing baselines, our \name achieves the best memory capabilities and the highest visual quality, strongly validating the efficacy of the proposed Implicit Memory Retrieval.
    } 
    \label{fig:qualitative}

    \vspace{-10pt}
\end{figure*}

\subsection{Experiment Settings}
\noindent\textbf{Implementation Details.}
We build \name on Wan2.2-TI2V-5B and generate 81-frame videos at $832\times480$ resolution. CaM~\cite{yu2025context} and HyDRA~\cite{chen2026hydra} are reimplemented under the same backbone, data, and output settings. We train for over 100,000 iterations using 16 NVIDIA H20 GPUs. Context clips are encoded as continuous sequences during training, whereas inference may use discontinuous context clips.

\noindent\textbf{Evaluation Settings.}
Our evaluation metrics include: (1) Fréchet Video Distance (FVD) for video quality assessment, and (2) PSNR, SSIM, and LPIPS for quantifying memory capability by measuring pixel-level and perceptual differences between frames. To evaluate our method, we hold out 560 videos of our SceneFly dataset, which contains diverse scenes, as the test set. By evaluating on this dataset featuring complex camera revisiting trajectories, we can effectively assess the model's generalization capabilities to novel trajectories. Furthermore, to evaluate the model's performance on real-world datasets, we also conduct tests on the SpatialVid dataset.

\noindent\textbf{Evaluation Methods.}
We compare our method with existing approaches across two primary tasks: (1) \textit{Video Extension}, which evaluates the generation of scene-consistent continuations; and (2) \textit{Scene Revisiting}, an essential capability for video world models that compares new and previously generated frames. To quantify the latter, we employ a round-trip camera trajectory (rotating $n^\circ$ and returning) to facilitate direct PSNR, SSIM, and LPIPS calculations.Furthermore, our method uniquely supports \textit{Camera Hard Cut} generation, synthesizing videos with discontinuous viewpoint transitions. By enforcing abrupt camera shifts for subsequent clips and independently VAE-encoding past contexts, this setting acts as a pure text-to-video (T2V) task devoid of local reference frames. This rigorously tests the model's spatial generalization and memory capabilities. As existing baselines lack this functionality, we exclusively present our method's results for this highly demanding task.

\subsection{Quantitative Results}

As shown in Table~\ref{tab:main}, the quantitative metrics demonstrate the clear advantages of our \name over existing approaches. Across the complex camera trajectories of SceneFly and the diverse scenes of SpatialVid, our method effectively retrieves and utilizes relevant context information, whereas other methods suffer from limited context accessibility. Despite its massive parameter count, Lingbot struggles with precise camera control and exhibits poor scene consistency, likely due to the restricted context length used during its training phase. Furthermore, the suboptimal performance of CaM and HyDRA stems from their reliance on explicit retrieval; the rigidity of their fixed heuristic rules fails to adapt to highly variable camera trajectories and environments. In contrast, our method employs implicit retrieval, delegating the entire retrieval process to the model. These superior results strongly validate the effectiveness of our approach. Furthermore, the significant performance degradation observed with naive downsampling (Ours-DS) strongly validates the necessity of our proposed Context Compression Network.

\subsection{Qualitative Results}

In Figure~\ref{fig:qualitative}, we show our approach against other state-of-the-art methods across two distinct tasks. As demonstrated, our method achieves significantly superior scene consistency compared to both CaM and HyDRA. Furthermore, despite the considerably larger parameter scale of LingBot-World, it struggles to maintain scene consistency, thereby highlighting the inherent limitations of its full-context strategy.

\begin{figure}[t]
    \centering
    \includegraphics[width=\linewidth]{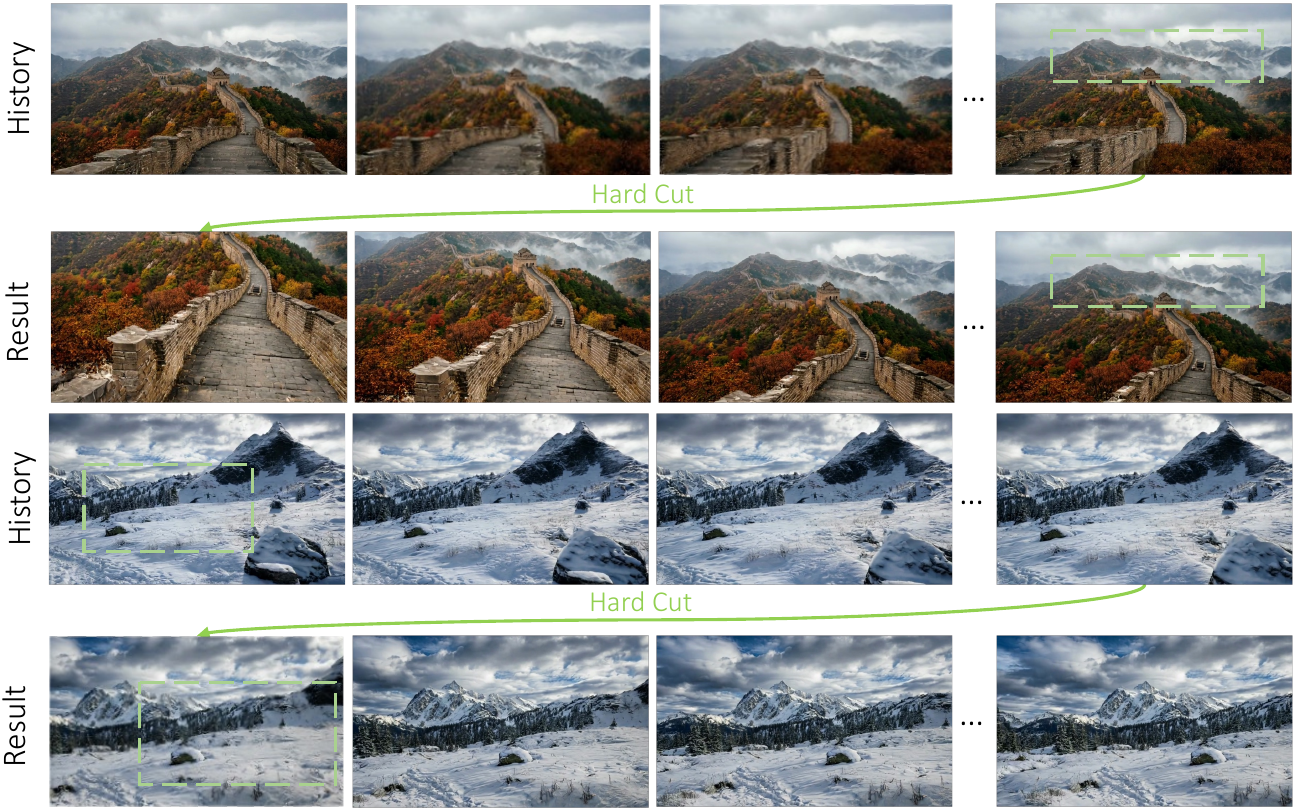}
\vspace{-10pt}
    \captionof{figure}{
    \textbf{Visual Result of Camera Hard Cut.}~Conditioned on the available context and discontinuous camera poses, our model successfully synthesizes videos with hard shot transitions without image reference. This compellingly demonstrates its memory retrieval capabilities and precise camera control.
    } 
    \label{fig:hard_cut}
\vspace{-10pt}
\end{figure}

In Figure~\ref{fig:hard_cut}, we present the qualitative results for the Camera Hard Cut task. As illustrated, our method successfully synthesizes scene-consistent videos with hard shot transitions conditioned on highly discontinuous camera poses. This compellingly demonstrates our model's exceptional memory retrieval capabilities and its generalization to complex camera trajectories.

\begin{figure*}[t]

    \centering
    \includegraphics[width=\linewidth]{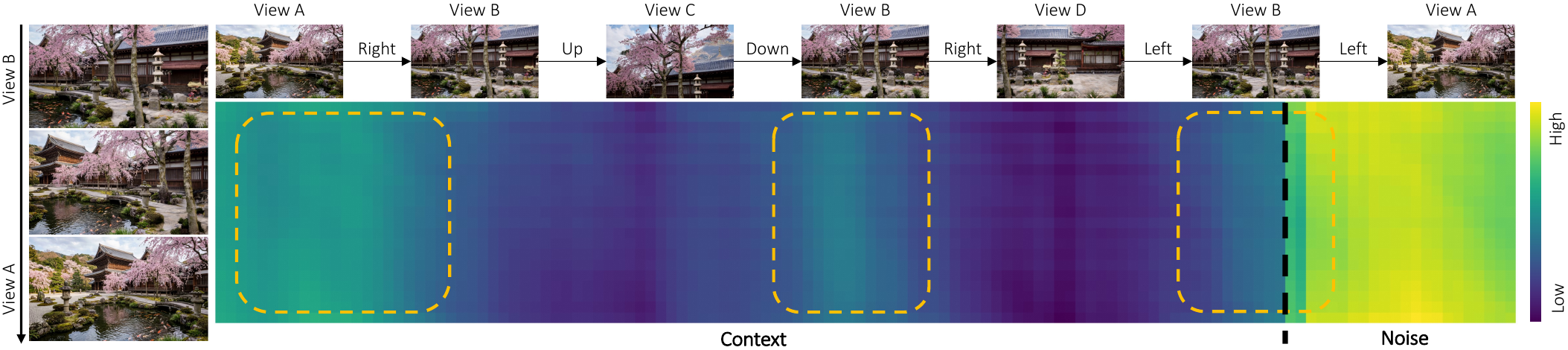}
    \vspace{-10pt}
    \caption{\textbf{Retrieval Attention visualization.}
    The first five columns correspond to context clips along A$\to$B$\to$C$\to$B$\to$D$\to$B, and column~6 is self-attention for the B$\to$A target. The model assigns stronger attention to context views that are closer to the target viewpoint: A$\to$B receives the highest context response, B$\leftrightarrow$C receives an intermediate response, and B$\leftrightarrow$D is most suppressed.}
    \label{fig:attention_viz}
\end{figure*}

\subsection{Ablation Study}

\noindent\textbf{Position Encoding.}
We conduct this ablation study specifically on the Camera Hard Cut task. As a significantly more challenging setting, it serves as a rigorous testbed to thoroughly evaluate the model's memory retrieval capabilities (Table~\ref{tab:ablation_pe}). Specifically, we evaluate the position encoding mechanism under three distinct configurations:
(1) Standard RoPE: Lacking viewpoint information, this baseline renders the model incapable of retrieving memory or controlling the camera, confirming the necessity of a dedicated encoding scheme for memory retrieval.
(2) Backbone Substitution: Directly replacing RoPE with relative pose encoding within the DiT backbone compromises the inherent capabilities of the base model, thereby justifying the need for an independent implicit retrieval module.
(3) Absolute Pose Encoding: Similar to CaM and HyDRA, this variant adds encoded absolute camera poses directly to the DiT latents. However, it fails to capture the relative geometric relationships among latents, struggling to maintain spatial awareness under varying camera trajectories. This finding strongly validates the superiority of our proposed relative pose encoding.
Finally, the slightly degraded FVD performance of our method can be attributed to the increased complexity of our relative pose encoding, which induces a more pronounced shift in the generated video distribution compared to the base model.

\begin{table}[t]
\centering
\small
\caption{\textbf{Ablation on Positional Encoding Strategy.}~Experimental results demonstrate the necessity of Relative Pose Encoding.}

\label{tab:ablation_pe}
\setlength{\tabcolsep}{4pt}
\begin{tabular}{lcccc}
\toprule
Method & PSNR$\uparrow$ & SSIM$\uparrow$ & LPIPS$\downarrow$ & FVD$\downarrow$ \\
\midrule
RoPE & 14.44 & 0.415 & 0.541 & 196.6 \\
w/o Retrival Attention & 13.49 & 0.416 & 0.572 & 445.1 \\
CaM \& HyDRA & 15.52 & 0.490 & 0.476 & \textbf{117.2} \\
Ours & \textbf{17.75} & \textbf{0.551} & \textbf{0.332} & 144.0 \\
\bottomrule
\end{tabular}

\end{table}

\noindent\textbf{Sink Strategy.}
Extending our method to long video generation introduces visual drift, driven by autoregressive error accumulation and the loss of fine-grained details from extreme context compression. To mitigate this, we prepend an independently encoded, uncompressed anchor frame to the noisy latents. Table~\ref{tab:ablation_sink} compares two anchor selection strategies:~(1)Tail Sink: Selects the context frame whose camera pose is closest to the target's final frame. While providing head-and-tail structural references, this frame is inherently model-generated, meaning the fundamental issue of error accumulation persists.
~(2)Initial Sink(Ours): Utilizes the user-provided initial image. As an error-free ground truth, this reference effectively arrests artifact accumulation, yielding significantly more stable and superior visual quality.

\begin{table}[t]
\centering
\small

\caption{\textbf{Ablation on Sink Strategy.}~When applying our method to long image-to-video generation, introducing the initial frame as a sink latent effectively mitigates the drift.}

\label{tab:ablation_sink}
\setlength{\tabcolsep}{4pt}
\begin{tabular}{lcccc}
\toprule
Method & PSNR$\uparrow$ & SSIM$\uparrow$ & LPIPS$\downarrow$ & FVD$\downarrow$ \\
\midrule
w/o sink & 15.97 & 0.472 & 0.436 & 153.2 \\
tail & 16.73 & 0.497 & 0.397 & 154.5 \\
Ours & \textbf{17.75} & \textbf{0.551} & \textbf{0.332} & \textbf{144.0} \\
\bottomrule
\end{tabular}
\end{table}

\subsection{Attention Logit Visualization}
\label{sec:appendix_attention}
\noindent\textbf{Setup.}
We visualize Retrieval Attention to examine whether the injected camera poses enable the model to retrieve memory according to viewpoint relevance. We construct a controlled round-trip trajectory over four views arranged in an L-shaped layout: A$\to$B$\to$C$\to$B$\to$D$\to$B$\to$A. View B lies to the right of A and strongly overlaps it; C lies above B and has only partial overlap with the A--B axis; D lies farther to the right of B and outside the field of view of A. The first five clips form the compressed context and serve as attention keys, while the final B$\to$A clip is the generation target and provides the queries. In Fig.~\ref{fig:attention_viz}, columns 1--5 show the attention logits assigned to the context clips, and column~6 shows target self-attention.

\noindent\textbf{Results.}
The responses follow the relative camera geometry. Clip~1 (A$\to$B) receives the strongest context response because it covers nearly the same region as the target. Clips~2--3 (B$\leftrightarrow$C) receive intermediate responses because they share evidence around B but deviate along the vertical branch. Clips~4--5 (B$\leftrightarrow$D) are most suppressed because they move away from the target viewing range. The ordering
\emph{self} $>$ \emph{A--B context} $>$ \emph{B--C context} $>$ \emph{B--D context}
shows that the attention strength varies consistently with viewpoint proximity. This pose-guided attention behavior realizes implicit memory retrieval without manually specifying which context frames should be selected.

\noindent\textbf{Analysis.}
This pattern reflects the central advantage of implicit retrieval. Long histories contain directly useful views, partial evidence, and distractors. Uniform aggregation can introduce irrelevant content, whereas hand-designed top-$k$ retrieval makes a discrete, irreversible decision before generation. Retrieval Attention instead assigns continuous, query-dependent weights: relative pose encoding provides pairwise geometry, while learned visual features determine generative utility. In particular, the intermediate B$\leftrightarrow$C responses show that the model can retain partial evidence rather than merely selecting or rejecting entire frames. This behavior is consistent with Sec.~\ref{sec:implicit}, where retrieval is learned inside the generator and can combine evidence across views.

%% file: section/conclusion.tex
\section{conclusion}
\label{conclusion}
In this work, we propose \name, an attention-driven implicit memory retrieval mechanism. This highly flexible paradigm eliminates the reliance on hand-crafted heuristics, enabling direct information retrieval across the entire context. Furthermore, our model maintains exceptional generation performance even when operating on highly compact contexts. Empowered by implicit memory retrieval, our approach successfully synthesizes scene-consistent long videos under complex camera trajectories, and is remarkably capable of generating hard-cut videos with discontinuous camera motions.

%% file: section/appendix.tex
\section{Effect of Compression Ratio}

We investigate the impact of different compression ratios on generation quality and efficiency. The compression ratio is denoted as $T{\times}H{\times}W$, representing the temporal, height, and width downsampling factors. Our default configuration uses $2{\times}4{\times}4$. We compare it against a more aggressive $4{\times}8{\times}8$ setting and a less aggressive $1{\times}2{\times}2$ setting. As shown in Table~\ref{tab:ablation_compression}, the $2{\times}4{\times}4$ ratio achieves the best trade-off between generation quality and efficiency. The $1{\times}2{\times}2$ ratio retains more spatial details but incurs higher latency and GPU memory overhead with marginal quality gains, while the $4{\times}8{\times}8$ ratio reduces cost at the expense of generation quality.

\begin{table}[h]
\centering
\small
\caption{\textbf{Effect of Compression Ratio.}~Ablation on compression ratios ($T{\times}H{\times}W$). Our default $2{\times}4{\times}4$ achieves the best quality--efficiency trade-off.}
\label{tab:ablation_compression}
\setlength{\tabcolsep}{3pt}
\begin{tabular}{lccccccc}
\toprule
Ratio & PSNR$\uparrow$ & SSIM$\uparrow$ & LPIPS$\downarrow$ & FVD$\downarrow$ & Latency $\downarrow$ & Mem. $\downarrow$ \\
\midrule
$1{\times}2{\times}2$ & \textbf{18.02} & \textbf{0.563} & \textbf{0.340} & 148.4 & 24.08s & 31.19GB \\
$2{\times}4{\times}4$ (default) & 17.75 & 0.551 & 0.332 & \textbf{144.0} & 3.32s & 17.83GB \\
$4{\times}8{\times}8$ & 17.42 & 0.541 & 0.324 & 141.1 & 1.98s & 17.34GB \\
\bottomrule
\end{tabular}
\vspace{-10pt}
\end{table}

\section{SceneFly Data Construction Pipeline}
\label{sec:appendix_scenefly}

\begin{figure*}[t]
    \centering
    \includegraphics[width=\linewidth]{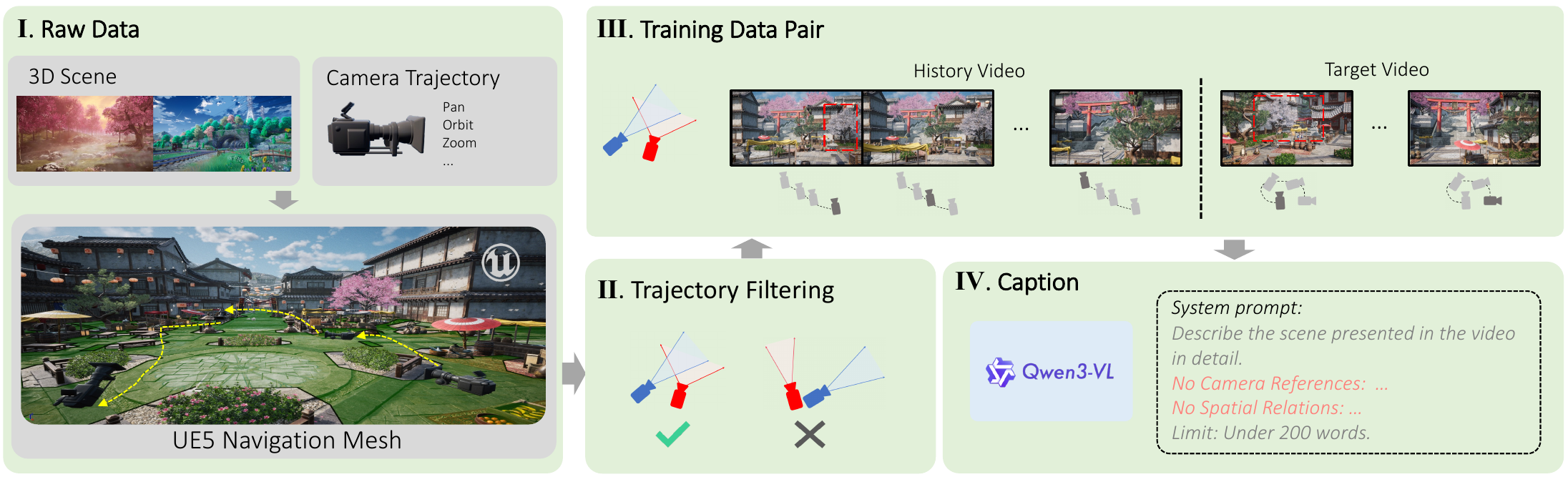}
    \vspace{-10pt}
    \caption{\textbf{Overview of the SceneFly Data Construction Pipeline.}~Our pipeline leverages the built-in navigation mesh system of UE5 to drive a camera that randomly explores each scene and renders long videos with camera trajectory. Training pairs are then constructed by sampling a target clip and assembling its context with a bias toward high FoV overlap. Finally, each target clip is precisely captioned by Qwen3-VL.}
    \label{fig:data_pipeline}
\end{figure*}

Training long-term scene memory requires videos that leave and revisit scene regions under diverse camera motions, together with accurate frame-level camera poses. Existing datasets provide limited scene diversity, monotonous trajectories, noisy pose estimates, or too few revisiting sequences to support controlled training and evaluation. SceneFly addresses these limitations with an automated Unreal Engine~5 pipeline that generates long exploration videos, exact camera annotations, target--context training pairs, and motion-agnostic captions. This section complements the overview in Figure~\ref{fig:data_pipeline} with implementation-level details of the SceneFly pipeline. We describe (i) the scene corpus and the atomic camera motion library, (ii) the Navigation-Mesh-based exploration algorithm that turns these primitives into long videos, (iii) the FoV-overlap-driven scheme that assembles target--context training pairs, and (iv) the caption protocol that decouples textual cues from camera motion.

\noindent\textbf{Scene Corpus and Atomic Motion Library.}
SceneFly is built upon $100$ high-fidelity Unreal Engine~5 scenes carefully curated from public marketplaces and in-house assets, spanning indoor environments (apartments, museums, offices), outdoor environments (urban streets, parks, natural landscapes), and stylized worlds (cartoonish, sci-fi, post-apocalyptic). The resulting dataset contains approximately 1,000 minutes of rendered video. To produce trajectories that mimic real user exploration, we predefine a library of \emph{atomic camera motions} that includes pure translations along all six cardinal directions, yaw/pitch rotations of varying magnitude, orbits around a focal point, dolly-in/out maneuvers, and idle hovering. Each atomic motion is parameterized by speed, amplitude, and duration, all of which are drawn from scene-aware ranges so that the resulting motion remains physically plausible (e.g., translation speeds are scaled by the local free-space radius of the scene). This atomic library is the source of the trajectory richness reported in the main paper.

\noindent\textbf{Automated Navigation Mesh and Free Exploration.}
For every scene, we automatically build a Navigation Mesh (NavMesh) by invoking UE5's recast-based navigation generator on a bounding volume that tightly encloses the scene's playable region. The generator rejects geometry that is non-walkable (steep slopes, occluded ceilings, inside-collider regions), yielding a polygonal mesh that exactly represents all camera-reachable areas. A virtual camera agent is then spawned on this NavMesh and explores it through a stochastic policy: at each decision step, the agent samples a new atomic motion from the library together with a NavMesh-feasible target waypoint, and the engine's path-finding routine produces a collision-free path to that waypoint. Whenever the agent risks leaving the mesh or colliding with thin geometry, the policy automatically resamples a new motion, ensuring uninterrupted long-horizon recordings. Each long video is then assembled from a sequence of $81$-frame atomic clips, with motion types resampled \emph{independently} between clips so that abrupt changes of viewing intent occur frequently. Per-frame camera extrinsics and intrinsics are logged in lock-step with the renderer, providing exact ground-truth poses.

\noindent\textbf{Training Pair Construction.}
Once a long video is collected, we transform it into multiple supervised training pairs through a graded FoV-overlap selection scheme. For each long video, we treat every atomic clip as a possible target $V_{tgt}$ and consider all other clips of the same video as context candidates. We project the camera frustum of the target's middle frame onto each candidate's frames and compute a frame-wise FoV intersection-over-union; the candidate's overlap score is the maximum value across its frames. Candidates are then partitioned into three buckets---\emph{strong overlap}, \emph{moderate overlap}, and \emph{disjoint}---and we sample context clips per pair from a mixture distribution that is intentionally skewed toward the strong and moderate buckets, while still preserving a small but non-trivial fraction of disjoint clips. This graded mixture is critical: the high-overlap clips supply the supervision signal for memory retrieval, the moderate clips drive the model to learn fine-grained spatial relevance, and the disjoint clips act as distractors that prevent the attention from collapsing onto trivially close frames. Finally, we randomize the number of selected context clips for each pair, so that a single long video produces training samples spanning a wide range of context lengths and difficulty levels.

\noindent\textbf{Caption Protocol.}
Captioning targets the \emph{target} clip only and is performed by Qwen3-VL with a structured prompt template. The template instructs the model to enumerate scene-level attributes---dominant objects, surface materials, lighting condition, time of day, and spatial layout---while explicitly forbidding any vocabulary that describes camera behavior (e.g., \emph{pan}, \emph{zoom}, \emph{dolly}, \emph{tracking shot}, \emph{handheld}). Each caption is then passed through a lightweight rule-based filter that detects residual motion-related tokens; flagged captions are regenerated with a stricter prompt until clean, ensuring that camera dynamics are conveyed exclusively through the trajectory input rather than leaking through text. Compared to off-the-shelf video captions, this protocol produces shorter, motion-agnostic descriptions that prevent the model from short-circuiting camera control with linguistic cues.

\begin{figure*}[t]
    \centering
    \includegraphics[width=\linewidth]{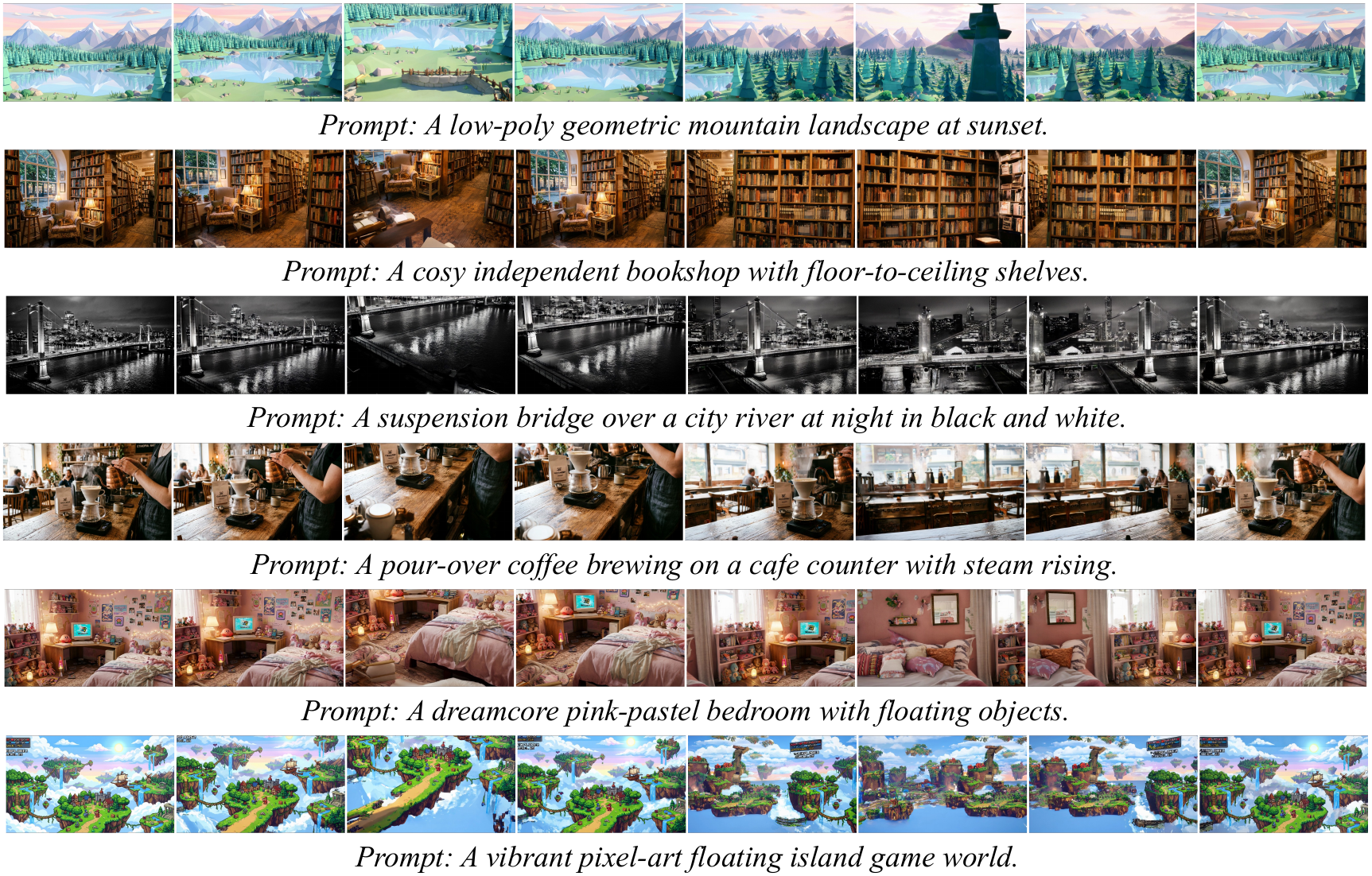}
    \caption{\textbf{Open-Domain Results (1).}}
    \label{fig:opendomain_1}
\end{figure*}

\begin{figure*}[t]
    \centering
    \includegraphics[width=\linewidth]{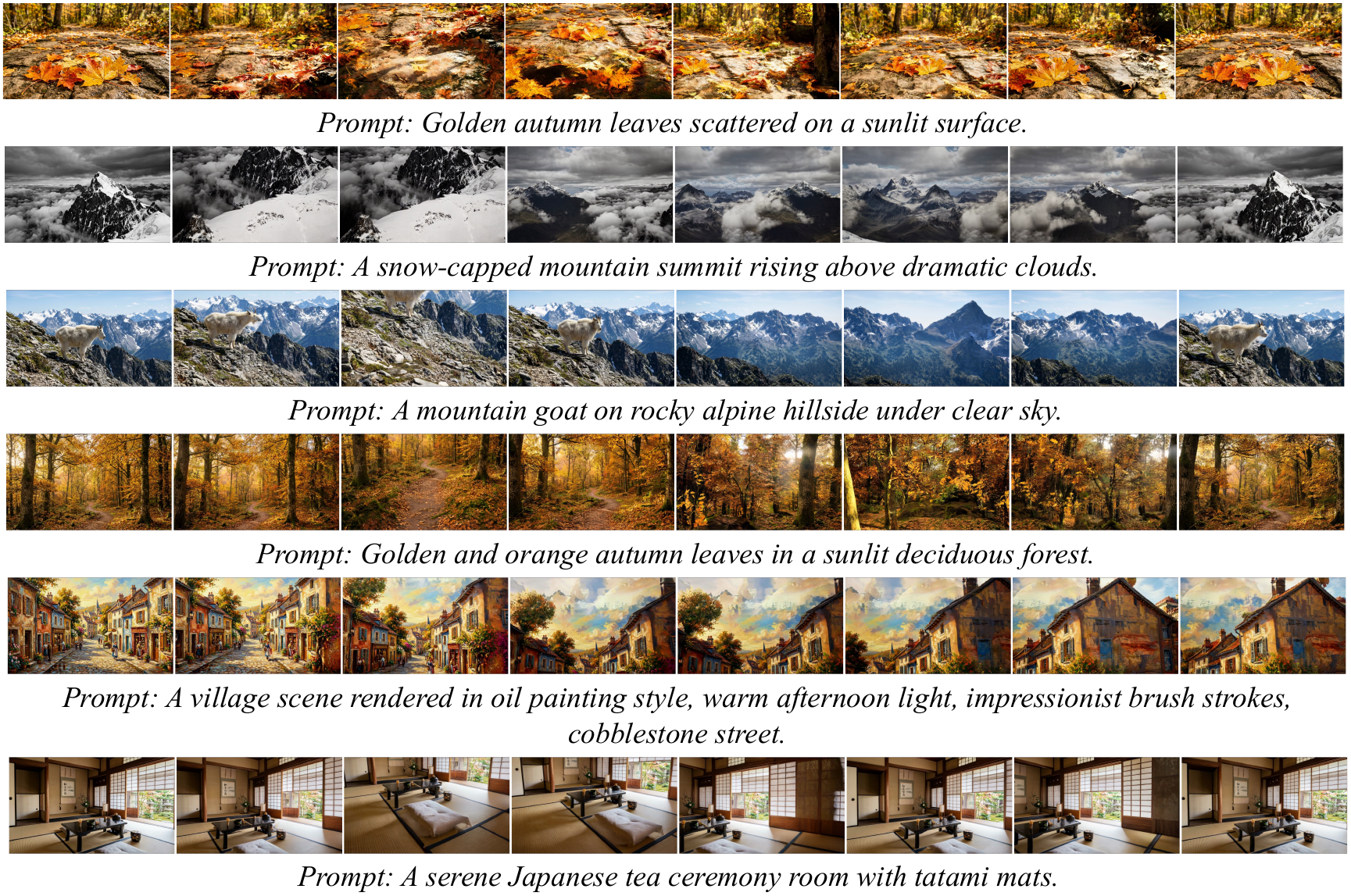}
    \caption{\textbf{Open-Domain Results (2).}}
    \label{fig:opendomain_2}
\end{figure*}

\section{Open-Domain Results}
\label{sec:appendix_opendomain}

Driven by our model's inherent capacity and the diversity of the SceneFly dataset, our method exhibits remarkable zero-shot generalization to open-domain scenarios. For validation, we curate stylistically diverse images from the internet---spanning real-world photographs, artistic renderings, and stylized illustrations---and use them as anchor frames for extended video synthesis. We use approximately round-trip trajectories: for most scenes, the camera's yaw and pitch are returned to zero near the end of the trajectory, while its translation remains unconstrained. Thus, the final view recovers the initial viewing orientation without necessarily returning to the original camera position, providing a challenging test of whether the model can preserve scene structure under combined rotational and translational changes. As shown in Figures~\ref{fig:opendomain_1} and \ref{fig:opendomain_2}, our method preserves robust memory retention and structural fidelity across a wide variety of open-domain environments.

\section{Visual Result of No History Content Overlap}
\label{sec:appendix_nooverlap}

\begin{figure}[H]
    \centering
    \includegraphics[width=\linewidth]{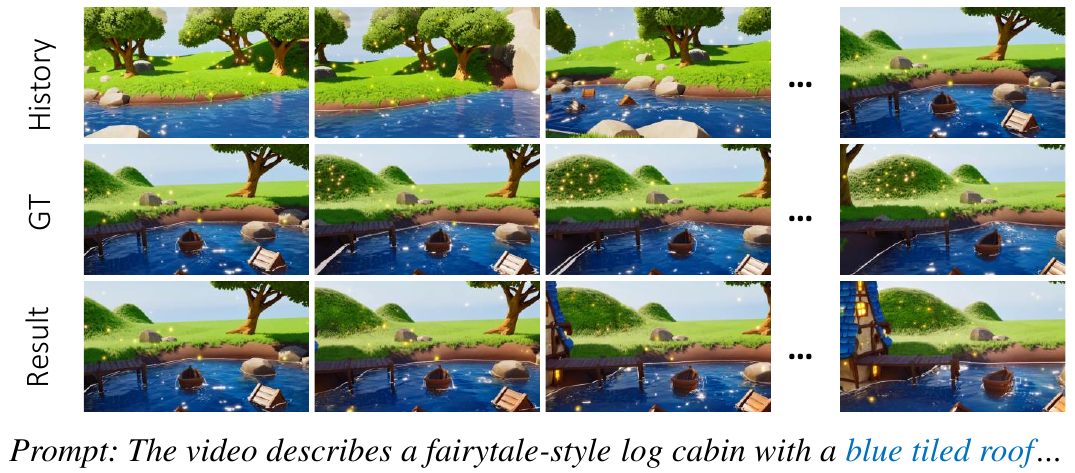}
    \caption{\textbf{Visual Result of No History Content Overlap.}~When the target trajectory shares no overlap with the historical context, the generated result follows the text instruction.}
    \label{fig:opendomain}
\end{figure}

We further investigate a corner case in which the target trajectory has \emph{no} spatial overlap with the available history---i.e., the target views something that has never been observed before. In this regime, no useful information can be retrieved from the context, so the model has to rely on the text prompt to synthesize plausible content. As shown in Figure~\ref{fig:opendomain}, our model gracefully falls back to following the textual instruction in this setting, generating coherent content consistent with the prompt while still respecting the specified camera trajectory.

\section{Limitations}
Our method has several limitations. First, SceneFly primarily emphasizes camera motion and scene revisiting but contains relatively few independently moving objects. As a result, \name is more effective at preserving static scene structure than modeling objects whose position or appearance changes over time. Historical observations of such objects may become outdated, and retrieving them without explicit motion reasoning can introduce temporal inconsistencies or recover an incorrect object state. Extending the training data with diverse dynamic scenes and incorporating motion-aware memory updates would help distinguish persistent scene content from time-varying states. Second, although retrieved history can anchor revisited regions, we do not introduce a dedicated anti-drift mechanism. Errors may therefore accumulate during long autoregressive rollouts, causing reduced sharpness, color shifts, or gradual deviations from the original scene appearance. Finally, \name does not operate in real time because each segment requires iterative diffusion sampling and attention over the compressed history. Step distillation, consistency-based training, and adaptive context computation are promising directions for reducing sampling and retrieval costs.

%% file: bibliography.bib
@String{Computer = "{IEEE} Computer" }

@String(CVPR= {IEEE Conf. Comput. Vis. Pattern Recog.})

@String(ICCV= {Int. Conf. Comput. Vis.})

@String(ICLR = {Int. Conf. Learn. Represent.})

@String(CVPR  = {CVPR})

@String(ICCV  = {ICCV})

@String(ICLR  = {ICLR})

@inproceedings{ren2025gen3c,
  title={{GEN3C}: 3D-Informed World-Consistent Video Generation with Precise Camera Control},
  author={Ren, Xuanchi and Shen, Tianchang and Huang, Jiahui and Ling, Huan and Lu, Yifan and Nimier-David, Merlin and M{\"u}ller, Thomas and Keller, Alexander and Fidler, Sanja and Gao, Jun},
  booktitle={Proceedings of the IEEE/CVF Conference on Computer Vision and Pattern Recognition},
  pages={6121--6132},
  year={2025}
}

@article{huang2026selfforcing,
  title={Self forcing: Bridging the train-test gap in autoregressive video diffusion},
  author={Huang, Xun and Li, Zhengqi and He, Guande and Zhou, Mingyuan and Shechtman, Eli},
  journal={Advances in Neural Information Processing Systems},
  volume={38},
  pages={167283--167308},
  year={2026}
}

@inproceedings{yin2025causvid,
    title={From Slow Bidirectional to Fast Autoregressive Video Diffusion Models},
    author={Yin, Tianwei and Zhang, Qiang and Zhang, Richard and Freeman, William T and Durand, Fredo and Shechtman, Eli and Huang, Xun},
    booktitle={CVPR},
    year={2025}
}

@inproceedings{bai2025recammaster,
  title={{ReCamMaster}: Camera-Controlled Generative Rendering from a Single Video},
  author={Bai, Jianhong and Xia, Menghan and Fu, Xiao and Wang, Xintao and Mu, Lianrui and Cao, Jinwen and Liu, Zuozhu and Hu, Haoji and Bai, Xiang and Wan, Pengfei and others},
  booktitle={Proceedings of the IEEE/CVF International Conference on Computer Vision},
  pages={14834--14844},
  year={2025}
}

@article{lingbot-world,
    title={Advancing Open-source World Models}, 
    author={Robbyant Team and Zelin Gao and Qiuyu Wang and Yanhong Zeng and Jiapeng Zhu and Ka Leong Cheng and Yixuan Li and Hanlin Wang and Yinghao Xu and Shuailei Ma and Yihang Chen and Jie Liu and Yansong Cheng and Yao Yao and Jiayi Zhu and Yihao Meng and Kecheng Zheng and Qingyan Bai and Jingye Chen and Zehong Shen and Yue Yu and Xing Zhu and Yujun Shen and Hao Ouyang},
    journal={arXiv preprint arXiv:2601.20540},
    year={2026}
}

@article{hong2025relic,
  title={Relic: Interactive video world model with long-horizon memory},
  author={Hong, Yicong and Mei, Yiqun and Ge, Chongjian and Xu, Yiran and Zhou, Yang and Bi, Sai and Hold-Geoffroy, Yannick and Roberts, Mike and Fisher, Matthew and Shechtman, Eli and others},
  journal={arXiv preprint arXiv:2512.04040},
  year={2025}
}

@inproceedings{yu2025context,
  title={Context as memory: Scene-consistent interactive long video generation with memory retrieval},
  author={Yu, Jiwen and Bai, Jianhong and Qin, Yiran and Liu, Quande and Wang, Xintao and Wan, Pengfei and Zhang, Di and Liu, Xihui},
  booktitle={Proceedings of the SIGGRAPH Asia 2025 Conference Papers},
  pages={1--11},
  year={2025}
}

@article{chen2026hydra,
  title={Out of Sight but Not Out of Mind: Hybrid Memory for Dynamic Video World Models},
  author={Chen, Kaijin and Liang, Dingkang and Zhou, Xin and Ding, Yikang and Liu, Xiaoqiang and Wan, Pengfei and Bai, Xiang},
  journal={arXiv preprint arXiv:2603.25716},
  year={2026}
}

@article{wan2025,
    title={Wan: Open and Advanced Large-Scale Video Generative Models}, 
    author={Team Wan and Ang Wang and Baole Ai and Bin Wen and Chaojie Mao and Chen-Wei Xie and Di Chen and Feiwu Yu and Haiming Zhao and Jianxiao Yang and Jianyuan Zeng and Jiayu Wang and Jingfeng Zhang and Jingren Zhou and Jinkai Wang and Jixuan Chen and Kai Zhu and Kang Zhao and Keyu Yan and Lianghua Huang and Mengyang Feng and Ningyi Zhang and Pandeng Li and Pingyu Wu and Ruihang Chu and Ruili Feng and Shiwei Zhang and Siyang Sun and Tao Fang and Tianxing Wang and Tianyi Gui and Tingyu Weng and Tong Shen and Wei Lin and Wei Wang and Wei Wang and Wenmeng Zhou and Wente Wang and Wenting Shen and Wenyuan Yu and Xianzhong Shi and Xiaoming Huang and Xin Xu and Yan Kou and Yangyu Lv and Yifei Li and Yijing Liu and Yiming Wang and Yingya Zhang and Yitong Huang and Yong Li and You Wu and Yu Liu and Yulin Pan and Yun Zheng and Yuntao Hong and Yupeng Shi and Yutong Feng and Zeyinzi Jiang and Zhen Han and Zhi-Fan Wu and Ziyu Liu},
    journal = {arXiv preprint arXiv:2503.20314},
    year={2025}
}

@article{kong2024hunyuanvideo,
  title={Hunyuanvideo: A systematic framework for large video generative models},
  author={Kong, Weijie and Tian, Qi and Zhang, Zijian and Min, Rox and Dai, Zuozhuo and Zhou, Jin and Xiong, Jiangfeng and Li, Xin and Wu, Bo and Zhang, Jianwei and others},
  journal={arXiv preprint arXiv:2412.03603},
  year={2024}
}

@article{wu2026pragmatic,
  title={A Pragmatic VLA Foundation Model},
  author={Wu, Wei and Lu, Fan and Wang, Yunnan and Yang, Shuai and Liu, Shi and Wang, Fangjing and Zhu, Qian and Sun, He and Wang, Yong and Ma, Shuailei and Ren, Yiyu and Zhang, Kejia and Yu, Hui and Zhao, Jingmei and Zhou, Shuai and Qiu, Zhenqi and Xiong, Houlong and Wang, Ziyu and Wang, Zechen and Cheng, Ran and Li, Yong-Lu and Huang, Yongtao and Zhu, Xing and Shen, Yujun and Zheng, Kecheng},
  journal={arXiv preprint arXiv:2601.18692},
  year={2026}
}

@article{hu2023gaia,
  title={Gaia-1: A generative world model for autonomous driving},
  author={Hu, Anthony and Russell, Lloyd and Yeo, Hudson and Murez, Zak and Fedoseev, George and Kendall, Alex and Shotton, Jamie and Corrado, Gianluca},
  journal={arXiv preprint arXiv:2309.17080},
  year={2023}
}

@inproceedings{yang2024cogvideox,
  title={CogVideoX: Text-to-Video Diffusion Models with An Expert Transformer},
  author={Yang, Zhuoyi and Teng, Jiayan and Zheng, Wendi and Ding, Ming and Huang, Shiyu and Xu, Jiazheng and Yang, Yuanming and Hong, Wenyi and Zhang, Xiaohan and Feng, Guanyu and others},
  booktitle={ICLR},
  year={2025}
}

@article{hong2022cogvideo,
  title={CogVideo: Large-scale Pretraining for Text-to-Video Generation via Transformers},
  author={Hong, Wenyi and Ding, Ming and Zheng, Wendi and Liu, Xinghan and Tang, Jie},
  journal={arXiv preprint arXiv:2205.15868},
  year={2022}
}

@misc{hunyuanvideo2025,
      title={HunyuanVideo 1.5 Technical Report}, 
      author={Wu, Bing and Zou, Chang and Li, Changlin and Huang, Duojun and Yang, Fang and Tan, Hao and Peng, Jack and Wu, Jianbing and Xiong, Jiangfeng and Jiang, Jie and others},
      year={2025},
      eprint={2511.18870},
      archivePrefix={arXiv},
      primaryClass={cs.CV},
      url={https://arxiv.org/abs/2511.18870}, 
}

@article{gao2024vista,
  title={Vista: A generalizable driving world model with high fidelity and versatile controllability},
  author={Gao, Shenyuan and Yang, Jiazhi and Chen, Li and Chitta, Kashyap and Qiu, Yihang and Geiger, Andreas and Zhang, Jun and Li, Hongyang},
  journal={Advances in Neural Information Processing Systems},
  volume={37},
  pages={91560--91596},
  year={2024}
}

@article{yang2023learning,
  title={Learning interactive real-world simulators},
  author={Yang, Sherry and Du, Yilun and Ghasemipour, Kamyar and Tompson, Jonathan and Kaelbling, Leslie and Schuurmans, Dale and Abbeel, Pieter},
  journal={arXiv preprint arXiv:2310.06114},
  year={2023}
}

@article{wang2025spatialvid,
  title={Spatialvid: A large-scale video dataset with spatial annotations},
  author={Wang, Jiahao and Yuan, Yufeng and Zheng, Rujie and Lin, Youtian and Gao, Jian and Chen, Lin-Zhuo and Bao, Yajie and Zhang, Yi and Zeng, Chang and Zhou, Yanxi and others},
  journal={arXiv preprint arXiv:2509.09676},
  year={2025}
}

@inproceedings{li2025prope,
 author = {Li, Ruilong and Yi, Brent and Liu, Junchen and Gao, Hang and Ma, Yi and Kanazawa, Angjoo},
 booktitle = {Advances in Neural Information Processing Systems},
 pages = {15984--16009},
 title = {Cameras as Relative Positional Encoding},
 volume = {38},
 year = {2025}
}

@inproceedings{zhang2025ucpe,
  title={Unified Camera Positional Encoding for Controlled Video Generation},
  author={Zhang, Cheng and Li, Boying and Wei, Meng and Cao, Yan-Pei and Gambardella, Camilo Cruz and Phung, Dinh and Cai, Jianfei},
  booktitle={CVPR},
  year={2026}
}

@article{brooks2024video,
  title={Video generation models as world simulators},
  author={Brooks, Tim and Peebles, Bill and Holmes, Connor and DePue, Will and Guo, Yufei and Jing, Li and Schnurr, David and Taylor, Joe and Luhman, Troy and Luhman, Eric and others},
  journal={OpenAI Blog},
  volume={1},
  pages={1},
  year={2024}
}

@inproceedings{peebles2023scalable,
  title={Scalable diffusion models with transformers},
  author={Peebles, William and Xie, Saining},
  booktitle={ICCV},
  pages={4195--4205},
  year={2023}
}

@inproceedings{guo2024animatediff,
  title={ANIMATEDIFF: ANIMATE YOUR PERSONALIZED TEXT-TO-IMAGE DIFFUSION MODELS WITHOUT SPECIFIC TUNING},
  author={Guo, Yuwei and Yang, Ceyuan and Rao, Anyi and Liang, Zhengyang and Wang, Yaohui and Qiao, Yu and Agrawala, Maneesh and Lin, Dahua and Dai, Bo},
  booktitle={ICLR},
  year={2024}
}

@article{liu2025free4d,
  title={Free4D: Tuning-free 4D Scene Generation with Spatial-Temporal Consistency},
  author={Liu, Tianqi and Huang, Zihao and Chen, Zhaoxi and Wang, Guangcong and Hu, Shoukang and Shen, Liao and Sun, Huiqiang and Cao, Zhiguo and Li, Wei and Liu, Ziwei},
  journal={arXiv preprint arXiv:2503.20785},
  year={2025}
}

@inproceedings{wang2024motionctrl,
  title={{MotionCtrl}: A Unified and Flexible Motion Controller for Video Generation},
  author={Wang, Zhouxia and Yuan, Ziyang and Wang, Xintao and Chen, Tianshui and Xia, Menghan and Luo, Ping and Shan, Ying},
  booktitle={ACM SIGGRAPH Conference Papers},
  pages={1--11},
  year={2024}
}

@inproceedings{he2025cameractrl,
  title={{CameraCtrl}: Enabling Camera Control for Video Diffusion Models},
  author={He, Hao and Xu, Yinghao and Guo, Yuwei and Wetzstein, Gordon and Dai, Bo and Li, Hongsheng and Yang, Ceyuan},
  booktitle={ICLR},
  year={2025}
}

@article{yu2025trajectorycrafter,
  title={{TrajectoryCrafter}: Redirecting Camera Trajectory for Monocular Videos via Diffusion Models},
  author={Yu, Mark and Hu, Wenbo and Xing, Jinbo and Shan, Ying},
  journal={arXiv preprint arXiv:2503.05638},
  year={2025}
}

@inproceedings{baisyncammaster,
  title={SynCamMaster: Synchronizing Multi-Camera Video Generation from Diverse Viewpoints},
  author={Bai, Jianhong and Xia, Menghan and Wang, Xintao and Yuan, Ziyang and Liu, Zuozhu and Hu, Haoji and Wan, Pengfei and ZHANG, Di},
  booktitle={ICLR},
  year={2025}
}

@inproceedings{lipman2023flow,
  title={Flow Matching for Generative Modeling},
  author={Lipman, Yaron and Chen, Ricky TQ and Ben-Hamu, Heli and Nickel, Maximilian and Le, Matt},
  booktitle={ICLR},
  year={2023}
}

@inproceedings{liu2023flow,
  title={Flow Straight and Fast: Learning to Generate and Transfer Data with Rectified Flow},
  author={Liu, Xingchao and Gong, Chengyue and Liu, Qiang},
  booktitle={ICLR},
  year={2023}
}

@article{zhang2026frame,
  title={Frame Context Packing and Drift Prevention in Next-Frame-Prediction Video Diffusion Models},
  author={Zhang, Lvmin and Cai, Shengqu and Li, Muyang and Wetzstein, Gordon and Agrawala, Maneesh},
  journal={Advances in Neural Information Processing Systems},
  volume={38},
  pages={30546--30566},
  year={2025}
}

@article{zhang2025pretraining,
  title={Pretraining Frame Preservation for Lightweight Autoregressive Video History Embedding},
  author={Zhang, Lvmin and Cai, Shengqu and Li, Muyang and Zeng, Chong and Lu, Beijia and Rao, Anyi and Han, Song and Wetzstein, Gordon and Agrawala, Maneesh},
  journal={arXiv preprint arXiv:2512.23851},
  year={2025}
}

@inproceedings{esser2024scaling,
  title={Scaling rectified flow transformers for high-resolution image synthesis},
  author={Esser, Patrick and Kulal, Sumith and Blattmann, Andreas and Entezari, Rahim and M{\"u}ller, Jonas and Saini, Harry and Levi, Yam and Lorenz, Dominik and Sauer, Axel and Boesel, Frederic and others},
  booktitle={Forty-first international conference on machine learning},
  year={2024}
}

@misc{li2025hunyuangamecrafthighdynamicinteractivegame,
    title={Hunyuan-GameCraft: High-dynamic Interactive Game Video Generation with Hybrid History Condition}, 
    author={Jiaqi Li and Junshu Tang and Zhiyong Xu and Longhuang Wu and Yuan Zhou and Shuai Shao and Tianbao Yu and Zhiguo Cao and Qinglin Lu},
    year={2025},
    url={https://arxiv.org/abs/2506.17201}, 
}

@article{hunyuanworld2025tencent,
    title={HunyuanWorld 1.0: Generating Immersive, Explorable, and Interactive 3D Worlds from Words or Pixels},
    author={Team HunyuanWorld},
    year={2025},
    journal={arXiv preprint}
}

@article{hyworld22026,
  title={HY-World 2.0: A Multi-Modal World Model for Reconstructing, Generating, and Simulating 3D Worlds},
  author={Team HY-World},
  journal={arXiv preprint},
  year={2026}
}

@article{Qwen3-VL,
      title={Qwen3-VL Technical Report}, 
      author={Shuai Bai and Yuxuan Cai and Ruizhe Chen and Keqin Chen and Xionghui Chen and Zesen Cheng and Lianghao Deng and Wei Ding and Chang Gao and Chunjiang Ge and Wenbin Ge and Zhifang Guo and Qidong Huang and Jie Huang and Fei Huang and Binyuan Hui and Shutong Jiang and Zhaohai Li and Mingsheng Li and Mei Li and Kaixin Li and Zicheng Lin and Junyang Lin and Xuejing Liu and Jiawei Liu and Chenglong Liu and Yang Liu and Dayiheng Liu and Shixuan Liu and Dunjie Lu and Ruilin Luo and Chenxu Lv and Rui Men and Lingchen Meng and Xuancheng Ren and Xingzhang Ren and Sibo Song and Yuchong Sun and Jun Tang and Jianhong Tu and Jianqiang Wan and Peng Wang and Pengfei Wang and Qiuyue Wang and Yuxuan Wang and Tianbao Xie and Yiheng Xu and Haiyang Xu and Jin Xu and Zhibo Yang and Mingkun Yang and Jianxin Yang and An Yang and Bowen Yu and Fei Zhang and Hang Zhang and Xi Zhang and Bo Zheng and Humen Zhong and Jingren Zhou and Fan Zhou and Jing Zhou and Yuanzhi Zhu and Ke Zhu},
	  journal={arXiv preprint arXiv:2511.21631},
      year={2025}
}

@inproceedings{henschel2024streamingt2v,
  title={Streaming{T2V}: Consistent, Dynamic, and Extendable Long Video Generation from Text},
  author={Henschel, Roberto and Khachatryan, Levon and Poghosyan, Hayk and Hayrapetyan, Daniil and Tadevosyan, Vahram and Wang, Zhangyang and Navasardyan, Shant and Shi, Humphrey},
  booktitle={CVPR},
  year={2025}
}

@inproceedings{chen2024seine,
  title={SEINE: Short-to-Long Video Diffusion Model for Generative Transition and Prediction},
  author={Chen, Xinyuan and Wang, Yaohui and Zhang, Lingjun and Zhuang, Shaobin and Ma, Xin and Yu, Jiashuo and Wang, Yali and Lin, Dahua and Qiao, Yu and Liu, Ziwei},
  booktitle={ICLR},
  year={2024}
}

@inproceedings{lu2024freelong,
  title={FreeLong: Training-Free Long Video Generation with SpectralBlend Temporal Attention},
  author={Lu, Yu and Liang, Yuanzhi and Zhu, Linchao and Yang, Yi},
  booktitle={NeurIPS},
  year={2024}
}

@article{yu2025viewcrafter,
  title={ViewCrafter: Taming Video Diffusion Models for High-fidelity Novel View Synthesis},
  author={Yu, Wangbo and Xing, Jinbo and Yuan, Li and Hu, Wenbo and Li, Xiaoyu and Huang, Zhipeng and Gao, Xiangjun and Wong, Tien-Tsin and Shan, Ying and Tian, Yonghong},
  journal={arXiv preprint arXiv:2409.02048},
  year={2024}
}

@article{zheng2024cami2v,
  title={CamI2V: Camera-Controlled Image-to-Video Diffusion Model},
  author={Zheng, Guangcong and Li, Teng and Jiang, Rui and Lu, Yehao and Wu, Tao and Li, Xi},
  journal={arXiv preprint arXiv:2410.15957},
  year={2024}
}

@inproceedings{liu2023zero123,
  title={Zero-1-to-3: Zero-shot One Image to 3D Object},
  author={Liu, Ruoshi and Wu, Rundi and Van Hoorick, Basile and Tokmakov, Pavel and Chatterjee, Carl and Seidenschwarz, Lorenzo and Vondrick, Carl and Arandjelovic, Relja and Caron, Mathilde and Schwing, Alexander G},
  booktitle={ICCV},
  year={2023}
}

@inproceedings{bruce2024genie,
  title={Genie: Generative Interactive Environments},
  author={Bruce, Jake and Dennis, Michael and Edwards, Ashley and Parker-Holder, Jack and Shi, Yuge and Hughes, Edward and Lai, Matthew and Mavalankar, Aditi and Steigerwald, Richie and Apps, Chris and others},
  booktitle={ICML},
  year={2024}
}

@inproceedings{alonso2024diamond,
  title={Diffusion for World Modeling: Visual Details Matter in Atari},
  author={Alonso, Eloi and Jelley, Adam and Micheli, Vincent and Kanervisto, Anssi and Storkey, Amos and Pearce, Tim and Fleuret, Fran{\c{c}}ois},
  booktitle={NeurIPS},
  year={2024}
}

@inproceedings{valevski2024gamengen,
  title={Diffusion Models Are Real-Time Game Engines},
  author={Valevski, Dani and Leviathan, Yaniv and Arar, Moab and Fruchter, Shlomi},
  booktitle={ICLR},
  year={2025}
}

@article{agarwal2025cosmos,
  title={Cosmos World Foundation Model Platform for Physical AI},
  author={{NVIDIA} and Agarwal, Niket and Ali, Arslan and Bala, Maciej and Balaji, Yogesh and Barker, Erik and Cai, Tiffany and Chattopadhyay, Prithvijit and Chen, Yongxin and Cui, Yin and Ding, Yifan and others},
  journal={arXiv preprint arXiv:2501.03575},
  year={2025}
}

@inproceedings{li2025vmem,
  title={{VMem}: Consistent Interactive Video Scene Generation with Surfel-Indexed View Memory},
  author={Li, Runjia and Torr, Philip and Vedaldi, Andrea and Jakab, Tomas},
  booktitle={ICCV},
  year={2025}
}

@article{wu2025longtermvwm,
  title={Video World Models with Long-term Spatial Memory},
  author={Wu, Tong and Yang, Shuai and Po, Ryan and Xu, Yinghao and Liu, Ziwei and Lin, Dahua and Wetzstein, Gordon},
  journal={arXiv preprint arXiv:2506.05284},
  year={2025}
}

@inproceedings{yu2025gamefactory,
  title={GameFactory: Creating New Games with Generative Interactive Videos},
  author={Yu, Jiwen and Qin, Yiran and Wang, Xintao and Wan, Pengfei and Zhang, Di and Liu, Xihui},
  booktitle={ICCV},
  year={2025}
}

@inproceedings{guo2025lct,
  title={Long Context Tuning for Video Generation},
  author={Guo, Yuwei and Yang, Ceyuan and Yang, Ziyan and Ma, Zhibei and Lin, Zhijie and Yang, Zhenheng and Lin, Dahua and Jiang, Lu},
  booktitle={ICCV},
  year={2025}
}

@article{chen2026contextforcing,
  title={Context Forcing: Consistent Autoregressive Video Generation with Long Context},
  author={Chen, Shuo and Wei, Cong and Sun, Sun and Nie, Ping and Zhou, Kai and Zhang, Ge and Yang, Ming-Hsuan and Chen, Wenhu},
  journal={arXiv preprint arXiv:2602.06028},
  year={2026}
}
